\documentclass[conference]{IEEEtran}
\IEEEoverridecommandlockouts

\usepackage{cite}
\usepackage{amsmath,amssymb,amsfonts}
\usepackage{algorithmic}
\usepackage{graphicx}
\usepackage{textcomp}
\usepackage{xcolor}
\usepackage{hyperref}
\usepackage{graphicx}
\usepackage{tablefootnote}
\usepackage{colortbl}
\usepackage{xcolor}
\usepackage{pgf}
\usepackage{url}
\usepackage{comment}
\usepackage{array}
\usepackage{caption}
\usepackage{multirow}
\usepackage[numbers]{natbib}

\newcommand{\colorcell}[2]{\cellcolor{green!#1}{#2}}
\def\BibTeX{{\mathbb{R}m B\kern-.05em{\sc i\kern-.025em b}\kern-.08em
    T\kern-.1667em\lower.7ex\hbox{E}\kern-.125emX}}
\begin{document}

\title{Style Ambiguity Loss Using CLIP\\
{\footnotesize }
}

\author{\IEEEauthorblockN{1\textsuperscript{st} James Baker}
\IEEEauthorblockA{Computer Science and Engineering \\
University of Maryland, Baltimore County\\
Catonsville, MD 21228 \\
jbaker15@umbc.edu}
}

\maketitle

\begin{abstract}
In this work, we explore using the style ambiguity training objective, originally used to approximate creativity, on a diffusion model. However, this objective requires the use of a pretrained classifier and a labeled dataset. We introduce new forms of style ambiguity loss that do not require training a new classifier or a labeled dataset. Instead of using a classifier, we generate centroids in the CLIP embedding space, and images are classified based on their relative distance to said centroids. We find the centroids via K-means clustering of an unlabeled dataset, as well as using text labels to generate CLIP embeddings, to be used as centroids. Code is available at \url{https://github.com/jamesBaker361/clipcreate}
\end{abstract}

\begin{IEEEkeywords}
CLIP, multimodal, diffusion
\end{IEEEkeywords}

\section{Introduction}
Much of generative imagery is trained to mimic the underlying dataset. Generating truly "creative" work was initiated by the Creative Adversarial Network \citep{ElgammalLEM17}, which used a style ambiguity loss to train a generator network to generate images that could not be classified as belonging to a particular style. However, every set of styles or concepts requires training a classifier before even training a model to generate images, which itself takes time and requires a labeled dataset, the collection and curation of which can be expensive. To circumvent these issues, we propose using a classification technique for style ambiguity that does not require any additional training and can be easily applied to any dataset, labeled or unlabeled. We demonstrate this by using reinforcement learning to finetune diffusion models with the various style ambiguity losses. Our contributions are as follows:
\begin{itemize}
  \item We developed versatile Text-based and K-Means-based creative style ambiguity losses that do not require training a separate style classifier on a labeled dataset, instead using CLIP for generating centroids.
  \item We find that diffusion models trained with our novel forms of style ambiguity loss score similar or higher on automated metrics of human appreciation to diffusion models trained without, and perform a user study to show that our novel models also score higher on components of creativity.
\end{itemize}

\section{Background}
\subsection{Creative Adversarial Network} \label{can}
A Generative Adversarial Network, or GAN \citep{goodfellow2014generative}, consists of two models, a generator and a discriminator. The generator generates samples from noise, and the discriminator detects if the samples are drawn from the real data or generated. During training, the generator is trained to trick the discriminator into classifying generated images as real, and the discriminator is trained to classify images correctly. Given a generator \(G: \mathbb{R}^{noise} \mathbb{R}ightarrow \mathbb{R}^{h \times w \times 3}\), a discriminator \(D: \mathbb{R}^{h \times w \times 3} \mathbb{R}ightarrow [0,1]\) real images \(x \in \mathbb{R}^{h \times w \times 3}\), and noise \(\mathcal{Z} \in \mathbb{R}^{noise}\), the objective is:
\[\underset{G}{min} \: \underset{D}{max} \: \mathbb{E}_{x}[log(D(x)]+\mathbb{E}_{\mathcal{Z}}[log(1-D(G(\mathcal{Z}))]\]

\citep{ElgammalLEM17} introduced the Creative Adversarial Network, or CAN, which was a DCGAN \citep{radford2016unsupervised} where the discriminator was also trained to classify real samples, minimizing the style classification loss. Given \(N\) classes of image (such as ukiyo-e, baroque, impressionism, etc.), the classification modules of the Discriminator \(D_C: \mathbb{R}^{h \times w \times 3} \mathbb{R}ightarrow \mathbb{R}^{N}\) that returns a probability distribution over the \(N_s\) style classes for an image and the real labels \(\ell \in \mathbb{R}^{N}\), the style classification loss was:
\[L_{SL}=\mathbb{E}_{x,\ell} [\textbf{CE}(D_C(x),\ell)]\]
Where \(\textbf{CE}\) is the cross entropy function.

The generator was also trained to generate samples that could not be easily classified as belonging to one class. This stylistic ambiguity is a proxy for creativity or novelty. Given a vector \(U \in \mathbb{R}^{N}\), where each entry \(u_1, u_2,,,u_N = \frac{1}{N}\), and the classification modules of the discriminator  \(D_C\)  the style ambiguity loss is:
\[L_{SA} = \mathbb{E}_{\mathcal{Z}} [\textbf{CE}(C(G(\mathcal{Z})),U)]\]
The discriminator was additionally trained to minimize \(L_{SL}\) and the generator was additionally trained to minimize \(L_{SA}\). 

\subsection{Diffusion}
A diffusion model aims to learn to iteratively remove the noise from a corrupted sample to restore the original. Starting with \(x_0\), the forward process \(q\) iteratively adds Gaussian noise to produce the noised version \(x_T\), using a noise schedule \(\beta_1...\beta_T\), which can be learned or manually set as a hyperparameter:
\[q(x_t | x_{t-1}) =\mathcal{N} (x_t ; \sqrt{1-\beta_t} x_{t-1}, \beta_t \mathbf{I})\]
More importantly, we also want to model the reverse process \(p\), that turns a noisy sample \(x_T\) back into \(x_0\), conditioned on some context \(c\). As \(x_T\) is the fully noised version, \(p(x_T|c)=\mathcal{N}(x_T; \mathbf{0}, \mathbf{I})\)
\[p_{\theta} (x_{t-1} | x_t,c)=\mathcal{N}(x_{t-1}; \mu_{\theta} (x_t,t,c), \Sigma_{\theta}(x_t,t,c))\]

Once the model has been trained, the reverse process, aka inference, generates a sample from noise \(x_T \sim \mathcal{N}(0,1)\). We use the DDIM technique \citep{song2022denoisingdiffusionimplicitmodels} for sampling. In this work, we the more popular variant, Stable Diffusion \citep{Rombach_2022_CVPR}, where \(x\) is replaced with a latent embedding \(\mathcal{E}(x)\), where \(\mathcal{E}: \mathbb{R}^{h \times w \times 3} \mathbb{R}ightarrow \mathbb{R}^{h_z \times w_z \times c_z} \) is a frozen autoencoder \citep{kingma2022autoencoding}, and \(h_z<h, w_z<w,c_z>3\).

\section{Methods}
\subsection{Denoising Diffusion Proximal Optimisation}

Introduced by \citep{black2023training}, Denoising Diffusion Proximal Optimisation, or DDPO, represents the reverse Diffusion Process as a Markov Decision Process \citep{bellman1957markovian}. A similar method was also pursued by \citep{fan2023dpok}. Reinforcement learning training was then applied to a pretrained diffusion model.  Following \citep{schulman2017proximal}, \citep{black2023training} also implemented clipping to protect the policy gradient from excessively large updates, and per prompt stat tracking to normalize rewards. We largely follow their method but use a different reward function. We fine-tune off of the pre-existing \textbf{stabilityai/stable-diffusion-2-base} checkpoint \citep{Rombach_2022_CVPR} downloaded from \url{https://huggingface.co/stabilityai/stable-diffusion-2-base}.

\subsection{Text Prompts}
DDPO does not require any new data, given that we are fine-tuning off of a pretrained checkpoint. However, each time we train the model, we must decide which text prompts to use to condition the generation of images. We used the set of (painting, drawing, art) as our text prompts.

\subsection{Datasets and Labels}
Training the CAN, of course, requires a labeled dataset. We used two different real datasets based off of WikiArt \citep{wikiartSalehE15}:
\begin{enumerate}
    \item \textbf{Full:} the WikiArt dataset as is. Consists of roughly 80k images.
    \item \textbf{Mediums: } We used BLIP \citep{li2022blip} to generate captions, and  then selected the 10 classes that had the highest fraction of their descriptions containing any of the words in the text prompt set (painting, drawing, art). We then used the images in WikiArt that belonged to those classes. Consists of roughly 20k images.
\end{enumerate}
This also meant we had two different sets of style class labels: \(L_{full}\), all 27 class labels, used with \textbf{Full} and  \(L_{med}\), the 10 labels used in \textbf{Mediums}.

\subsection{DDPO Reward Function}\label{reward}
In the original DDPO paper, the authors used four different reward functions for four different tasks. In this paper, we use the reward model based on stylistic ambiguity, combined with a reward for utility. Given a pretrained CLIP \citep{radford2021clip} model, that can return a similarity score for each image-text pair: \(CLIP: \mathbb{R}^{text} \times \mathbb{R}^{h \times w \times 3} \mathbb{R}ightarrow \mathbb{R} \), image \(x_0 \in \mathbb{R}^{h \times w \times 3}\) generated with text prompt \(s\), cross entropy \(\textbf{CE}\), uniform distribution \(U \in \mathbb{R}^{N}\) and a classifier \(C: \mathbb{R}^{h \times w \times 3} \mathbb{R}ightarrow \mathbb{R}^{N}\) we want to maximize:
\[R(x_0) =  \lambda_{novelty} \textbf{CE}(C(x_0),U)+ \lambda_{utility} CLIP(s,x_0) \]
The first term on the left side of the equation represents style ambiguity loss, and the second term maintains alignment between text and image, essentially keeping the model from straying "too far" from the text prompt; these terms approximate novelty and utility, respectively. We actually have \textit{multiple} choices of classifier, which we discuss. The \textbf{Discriminator Classifier} is what was used for the CAN, and the \textbf{Text-Based Style Classifier} and \textbf{K-Means Image Based Classifer} have never been used for style ambiguity loss before, and our contribution is to introduce them and show their merits. Many other works \citep{wang2022clip2ganbridgingtextlatent,ye2023ipadapter} have incorporated CLIP embeddings as a condition to generate images, but we only use the CLIP classifier to generate a reward. 

\subsubsection{Discriminator Classifier}
We can use the classification module of the CAN discriminator as the classifier in the reward function, setting \(C=D_C\). This is the baseline against which we are comparing the other two types of classifier with. This discriminator was trained at image resolution 512, with batch size 128. Given that we had two datasets (\textbf{Full} and \textbf{Mediums}), we actually had to train two of these discriminators, one trained for each. 

\subsubsection{Text-Based Style Classifier}
For each generated image \(x_0\), for each class name \(s_i, 1 \le i \le N_s\) in the style class label set we want to use, we find \(CLIP(s_i,x_0)\). We can then create a vector \((CLIP(s_1,x_0), CLIP(s_2,x_0),,,CLIP(s_{N_s},x_0))\) and then use softmax to normalize the vector and define the result as \(TextC(x_0)\). Formally:
\[TextC(x_0)=\textbf{softmax}((CLIP(s_1,x_0),,,CLIP(s_{N_s},x_0))\]
Then we set \(C=TextC\). Given the two sets of labels, \(L_{full}\) and \(L_{med}\), we had two different types of Text-Based classifier.
\subsubsection{K-Means Image Based Classifiers}
Alternatively, we can embed the images (In our case, we embedded them into the CLIP embedding space \(\in \mathbb{R}^{768}\)) and perform k-means clustering to generate k centers. Given a CLIP Embedder \(E: \mathbb{R}^{h \times w \times 3} \mathbb{R}ightarrow \mathbb{R}^{768}\) mapping images to embeddings, and the k centers \(c_1, c_2,,,c_k\) we can create a vector \((\frac{1}{||E(x_0)-c_1||},\frac{1}{||E(x_0)-c_2||},,,,\frac{1}{||E(x_0)-c_k||}\) and then use softmax to normalize the vector and define the result as \(KM(x_0))\). Formally:
\[KM(x_0)=\textbf{softmax}(\frac{1}{||E(x_0)-c_1||},,,\frac{1}{||E(x_0)-c_k||})\]
Then we set \(C=KM\). Given two datasets (\textbf{Full} and \textbf{Mediums}), we had two sets of centers.

\section{Results}
Given three types of classifier and two datasets, we had six models. Each model was trained for 100 epochs on a single NVIDIA A100 Tensor Core GPU, elapsing roughly 14 hours each. Note that for methods \textit{text-full} and \textit{text-med}, we don't use the data in the data for the classifier per se; however, we use the style class labels, which are unique to the datasets in question. \textit{text-full} uses \(L_{full}\) and \textit{text-med} uses \(L_{mediums}\). We generated all images with width and height = 512.

\subsection{Automated Evaluation} 
For each dataset, for each choice of classifier, we made an evaluation dataset of 200 images. We used the exact same prompts and random seeds for each. I.e. if the nth image generated by \textit{disc-full} used prompt "painting" and random seed = \(z \in \mathbb{Z}\), so would the nth image generated by \textit{text-full}, \textit{kmeans-full}, etc. for prompts, we had 24 prompts, each generated by concatenating one of ["painting","art","drawing"] with one of [" of a man ", " of a woman "," of nature "," of an animal ", " of a city"," "," of a building "," of a child "]
We used two scoring metrics to score the models. Both of these metrics are metrics of human appreciation:
\begin{itemize}
    \item \textbf{AVA Score: (AVA)} The AVA model was trained on the AVA dataset \citep{avamurray} of images and average rankings by human subjects, in order to learn to approximate human preferences given an image. We used the CLIP model weights from the \textbf{clip-vit-large-patch14} checkpoint and the Multi-Layer Perceptron weights downloaded from \url{https://huggingface.co/trl-lib/ddpo-aesthetic-predictor}.
    \item \textbf{Image Reward: (IR)} The image reward model \citep{xu2023imagereward} was trained to score images given their text description based on a dataset of images and human rankings. We used the \textbf{image-reward} python library found at \url{https://github.com/THUDM/ImageReward/tree/main}.

\end{itemize}

\begin{table}[h]
    \centering
    \begin{tabular}{|c|c | c|}
    \hline
    Model & AVA  & IR   \\
    \hline
\(disc-full\)  &  \colorcell{58}{4.96 (0.5)} &  \colorcell{35}{-1.26 (1.04)} \\
\(clip-full\)  &  \colorcell{58}{4.99 (0.46)} &  \colorcell{34}{-1.59 (0.63)} \\
\(kmeans-full\)  &  \colorcell{100}{5.71 (0.42)} &  \colorcell{41}{0.03 (1.35)} \\
\hline
\(disc-med\)  &  \colorcell{63}{5.12 (0.4)} &  \colorcell{34}{-1.55 (0.59)} \\
\(clip-med\)  &  \colorcell{72}{5.44 (0.41)} &  \colorcell{52}{0.91 (0.89)} \\
\(kmeans-med\)  &  \colorcell{61}{5.09 (0.67)} &  \colorcell{50}{0.77 (0.87)} \\
\hline
    \end{tabular}
    \caption{Aesthetic Scores, Diffusion Models}
    \label{tab:aesthetic}
\end{table}

Results of our experiments  are shown in Table \ref{tab:aesthetic} Each cell contains the mean (and standard deviation). For both datasets, for both metrics, either a text model or kmeans model scored higher than the baseline discriminator model. This shows that our discriminator-free approaches in do not come at the cost of lower quality, and in fact are likely to produce \textit{better} images than the discriminator baseline.

\subsection{User Study}

We also conducted a user study to evaluate a subset of the models. The goal of this study was to evaluate each model for human appreciation, as we had done with IR and AVA, as well as novelty, surprise, ambiguity and complexity.  
For each model we \textit{did} evaluate in this user study, we generated five images with the same seeds. Then, for each image, users to gauge 5 metrics, by asking them each a question, as was done in \citep{ElgammalLEM17}. These 5 metrics were based on the components of aesthetic arousal \citep{berlyne1971aesthetics}, that being the level of alertness or excitation that an image provokes in a viewer. "Too much" arousal can actually have a negative effect on subjective appreciation. The metrics and their associated questions are as follows:
\begin{enumerate}
    \item \textbf{Appreciation (App):} How much do you like this image: 1-extremely dislike, 2-dislike, 3-Neutral, 4-like, 5-extremely like.
    \item \textbf{Novelty (Nov):} Rate the novelty of the image: 1-extremely not novel, 2-some how not novel, 3-neutral, 4 somehow novel, 5-extremely novel.
    \item \textbf{Surprise (Sur):}  Do you find the image surprising: 1-extremely not surprising, 2-some how not surprising, 3-neutral, 4-some how surprising, 5-extremely surprising.  
    \item \textbf{Ambiguity (Amb):}  Rate the ambiguity of the image. I find this image: 1-extremely not ambiguous, 2-some how not ambiguous, 3-neutral, 4-some how ambiguous, 5-extremely ambiguous.  
    \item \textbf{Complexity (Com):} Rate the complexity of the image. I find this image: 1-extremely simple, 2-some how simple, 3-neutral, 4-somehow complex, 5-extremely complex  
\end{enumerate}
We used \url{prolific.com} to find users, who were then redirected to a google forms survey. Each user was paid 10.00 dollars. Median time to complete the survey was 12 minutes and 28 seconds.

\begin{table}[h]
    \centering
    \renewcommand{\arraystretch}{1.2}
    \begin{tabular}{|lccccc|}
        \hline
        {} & App & Nov & Sur & Amb & Com \\
        \hline
        disc-full & \cellcolor{green!61}3.048 & \cellcolor{green!39}2.784 & \cellcolor{green!24}2.576 & \cellcolor{green!50}2.920 & \cellcolor{green!60}3.040 \\
        clip-full & \cellcolor{green!75}3.376 & \cellcolor{green!50}2.920 & \cellcolor{green!33}2.728 & \cellcolor{green!28}2.656 & \cellcolor{green!70}3.224 \\
        kmeans-full & \cellcolor{green!65}3.104 & \cellcolor{green!37}2.760 & \cellcolor{green!28}2.656 & \cellcolor{green!42}2.808 & \cellcolor{green!72}3.248 \\
        \hline
        disc-med & \cellcolor{green!17}2.600 & \cellcolor{green!33}2.728 & \cellcolor{green!11}2.520 & \cellcolor{green!57}3.144 & \cellcolor{green!62}3.064 \\
        clip-med & \cellcolor{green!48}2.936 & \cellcolor{green!55}3.032 & \cellcolor{green!42}2.808 & \cellcolor{green!39}2.784 & \cellcolor{green!58}3.104 \\
        kmeans-med & \cellcolor{green!68}3.216 & \cellcolor{green!80}3.328 & \cellcolor{green!50}2.960 & \cellcolor{green!85}3.448 & \cellcolor{green!100}3.728 \\
        
        \hline
    \end{tabular}
    \caption{User Study Results}
    \label{tab:green_shading}
\end{table}
Results are shown in table \ref{tab:green_shading}. We see here that \textit{kmeans-med} outperforms all of the other models on all of the metrics. For both datasets, the lowest Appreciation score was attained by the baseline discriminator model. For the novelty metric, which is a proxy for the "creative" aspect of creativity, both discriminator-free methods outperform the discriminator using the mediums datasets. 

\subsection{Visual Results}
\begin{figure}
    \centering
    \includegraphics[scale=0.3]{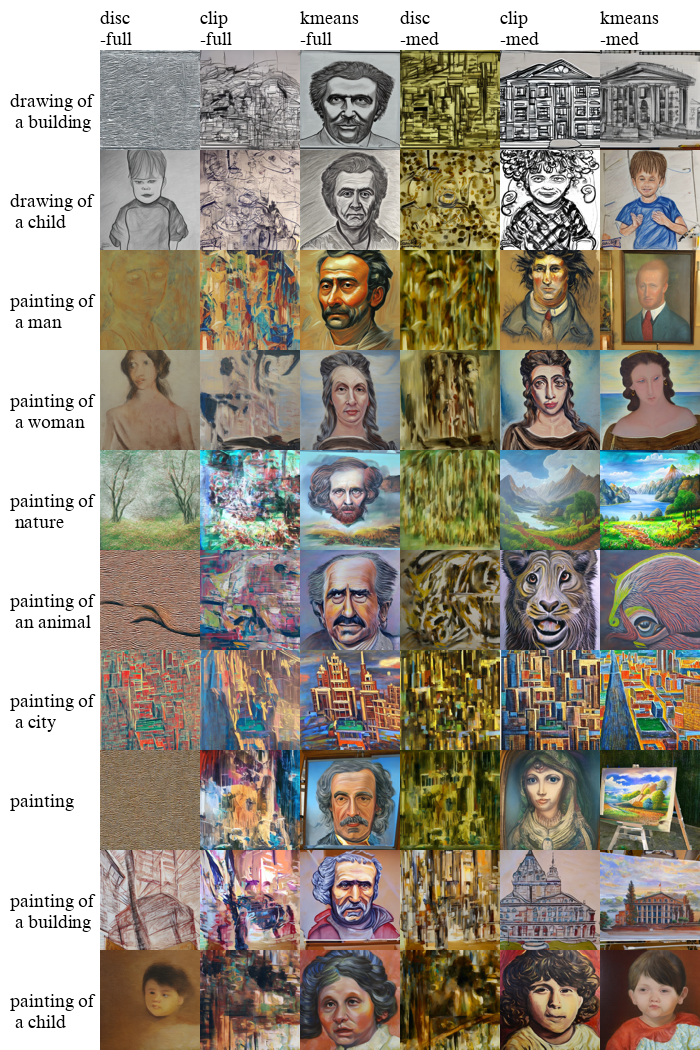}
    \caption{Image Comparisons}
    \label{fig:app1}
\end{figure}

We also provide a few visual results in figure \ref{fig:app1}. Each image in each row was generated with the same prompt and random seed. The leftmost column has the prompt used to generate each picture in each row, and each column is labeled with the model used to generate the images in said column.

Both discriminator model, as well as \textit{text-full} generate blurry, abstract patterns. While those are certainly interesting, they do not match the prompt. However, both \textit{text-med} and \textit{kmeans-med} demonstrate prompt alignment and interesting visuals.

\newpage

\section{Conclusion}
In this paper, we introduce versatile CLIP-based and K-Means-based creative style ambiguity losses, which eliminate the need for training a separate style classifier on labeled datasets. We have demonstrated that these new style ambiguity losses are just as good, oftentimes better, than the baseline discriminator based style ambiguity loss, as shown by automated metrics and user studies. This method can be applied to different image datasets or even different modalities like audio or video. Using the K-Means Classifier, or any unsupervised clustering method, for style ambiguity loss is particularly adept at this, given there is no need for \textit{any} labels.

\bibliographystyle{IEEEtran}
\bibliography{ieee}

\begin{thebibliography}{10}
\providecommand{\url}[1]{#1}
\csname url@samestyle\endcsname
\providecommand{\newblock}{\relax}
\providecommand{\bibinfo}[2]{#2}
\providecommand{\BIBentrySTDinterwordspacing}{\spaceskip=0pt\relax}
\providecommand{\BIBentryALTinterwordstretchfactor}{4}
\providecommand{\BIBentryALTinterwordspacing}{\spaceskip=\fontdimen2\font plus
\BIBentryALTinterwordstretchfactor\fontdimen3\font minus \fontdimen4\font\relax}
\providecommand{\BIBforeignlanguage}[2]{{%
\expandafter\ifx\csname l@#1\endcsname\relax
\typeout{** WARNING: IEEEtran.bst: No hyphenation pattern has been}%
\typeout{** loaded for the language `#1'. Using the pattern for}%
\typeout{** the default language instead.}%
\else
\language=\csname l@#1\endcsname
\fi
#2}}
\providecommand{\BIBdecl}{\relax}
\BIBdecl

\bibitem{ElgammalLEM17}
\BIBentryALTinterwordspacing
A.~M. Elgammal, B.~Liu, M.~Elhoseiny, and M.~Mazzone, ``{CAN:} creative adversarial networks, generating "art" by learning about styles and deviating from style norms,'' \emph{CoRR}, vol. abs/1706.07068, 2017. [Online]. Available: \url{http://arxiv.org/abs/1706.07068}
\BIBentrySTDinterwordspacing

\bibitem{goodfellow2014generative}
I.~J. Goodfellow, J.~Pouget-Abadie, M.~Mirza, B.~Xu, D.~Warde-Farley, S.~Ozair, A.~Courville, and Y.~Bengio, ``Generative adversarial networks,'' 2014.

\bibitem{radford2016unsupervised}
A.~Radford, L.~Metz, and S.~Chintala, ``Unsupervised representation learning with deep convolutional generative adversarial networks,'' 2016.

\bibitem{song2022denoisingdiffusionimplicitmodels}
\BIBentryALTinterwordspacing
J.~Song, C.~Meng, and S.~Ermon, ``Denoising diffusion implicit models,'' 2022. [Online]. Available: \url{https://arxiv.org/abs/2010.02502}
\BIBentrySTDinterwordspacing

\bibitem{Rombach_2022_CVPR}
R.~Rombach, A.~Blattmann, D.~Lorenz, P.~Esser, and B.~Ommer, ``High-resolution image synthesis with latent diffusion models,'' in \emph{Proceedings of the IEEE/CVF Conference on Computer Vision and Pattern Recognition (CVPR)}, June 2022, pp. 10\,684--10\,695.

\bibitem{kingma2022autoencoding}
D.~P. Kingma and M.~Welling, ``Auto-encoding variational bayes,'' 2022.

\bibitem{black2023training}
K.~Black, M.~Janner, Y.~Du, I.~Kostrikov, and S.~Levine, ``Training diffusion models with reinforcement learning,'' 2023.

\bibitem{bellman1957markovian}
\BIBentryALTinterwordspacing
R.~Bellman, ``A markovian decision process,'' \emph{Journal of Mathematics and Mechanics}, vol.~6, no.~5, pp. 679--684, 1957. [Online]. Available: \url{http://www.jstor.org/stable/24900506}
\BIBentrySTDinterwordspacing

\bibitem{fan2023dpok}
Y.~Fan, O.~Watkins, Y.~Du, H.~Liu, M.~Ryu, C.~Boutilier, P.~Abbeel, M.~Ghavamzadeh, K.~Lee, and K.~Lee, ``Dpok: Reinforcement learning for fine-tuning text-to-image diffusion models,'' 2023.

\bibitem{schulman2017proximal}
J.~Schulman, F.~Wolski, P.~Dhariwal, A.~Radford, and O.~Klimov, ``Proximal policy optimization algorithms,'' 2017.

\bibitem{wikiartSalehE15}
\BIBentryALTinterwordspacing
B.~Saleh and A.~M. Elgammal, ``Large-scale classification of fine-art paintings: Learning the right metric on the right feature,'' \emph{CoRR}, vol. abs/1505.00855, 2015. [Online]. Available: \url{http://arxiv.org/abs/1505.00855}
\BIBentrySTDinterwordspacing

\bibitem{li2022blip}
J.~Li, D.~Li, C.~Xiong, and S.~Hoi, ``Blip: Bootstrapping language-image pre-training for unified vision-language understanding and generation,'' in \emph{ICML}, 2022.

\bibitem{radford2021clip}
A.~Radford, J.~W. Kim, C.~Hallacy, A.~Ramesh, G.~Goh, S.~Agarwal, G.~Sastry, A.~Askell, P.~Mishkin, J.~Clark, G.~Krueger, and I.~Sutskever, ``Learning transferable visual models from natural language supervision,'' 2021.

\bibitem{wang2022clip2ganbridgingtextlatent}
\BIBentryALTinterwordspacing
Y.~Wang, W.~Zhou, J.~Bao, W.~Wang, L.~Li, and H.~Li, ``Clip2gan: Towards bridging text with the latent space of gans,'' 2022. [Online]. Available: \url{https://arxiv.org/abs/2211.15045}
\BIBentrySTDinterwordspacing

\bibitem{ye2023ipadapter}
H.~Ye, J.~Zhang, S.~Liu, X.~Han, and W.~Yang, ``Ip-adapter: Text compatible image prompt adapter for text-to-image diffusion models,'' 2023.

\bibitem{avamurray}
\BIBentryALTinterwordspacing
N.~Murray, L.~Marchesotti, and F.~Perronnin, ``Ava: A large-scale database for aesthetic visual analysis,'' 2016. [Online]. Available: \url{https://github.com/imfing/ava_downloader}
\BIBentrySTDinterwordspacing

\bibitem{xu2023imagereward}
J.~Xu, X.~Liu, Y.~Wu, Y.~Tong, Q.~Li, M.~Ding, J.~Tang, and Y.~Dong, ``Imagereward: Learning and evaluating human preferences for text-to-image generation,'' 2023.

\bibitem{berlyne1971aesthetics}
D.~E. Berlyne, \emph{Aesthetics and Psychobiology}.\hskip 1em plus 0.5em minus 0.4em\relax New York: Appleton-Century-Crofts, 1971.

\end{thebibliography}

\end{document}